\pdfoutput=1
\IfFileExists{./prepreamble-amspreprint.sty}{\RequirePackage[packages,theorems,changes]{./prepreamble-amspreprint}}{}

\makeatletter
\IfFileExists{./scoop-latex/scoop-packages.sty}{\providecommand*{\input@path}{}\edef\input@path{{./scoop-latex/}\input@path}}{}
\@namedef{ver@minted.sty}{}
\@namedef{opt@minted.sty}{}
\makeatother
\PassOptionsToPackage{style=authoryear-comp,backend=biber}{biblatex}
\PassOptionsToPackage{commentmarkup = footnote}{changes}    

\documentclass[english]{amsart}

\RequirePackage{biblatex}
\RequirePackage{ltxcmds}
\IfFileExists{./preamble-amspreprint.sty}{\RequirePackage[packages,theorems,changes]{./preamble-amspreprint}}{}

\usepackage{scoop-local}

\IfFileExists{./postpreamble-amspreprint.sty}{\RequirePackage[packages,theorems,changes]{./postpreamble-amspreprint}}{}

\makeatletter
\@ifpackageloaded{changes}{
\definechangesauthor[name = {Roland Herzog}, color = {red!80!black}]{RH}
\definechangesauthor[name = {Frederik Köhne}, color = {blue!80!black}]{FK}
\definechangesauthor[name = {Leonie Kreis}, color = {orange!80!black}]{LK}
\definechangesauthor[name = {Anton Schiela}, color = {green!70!black}]{AS}
}{}
\makeatother        

\makeatletter
\@ifpackageloaded{hyperref}{%
	\hypersetup{
		pdftitle = {Frobenius-Type Norms and Inner Products of Matrices and Linear Maps with Applications to Neural Network Training},
		pdfauthor = {Roland Herzog, Frederik Köhne, Leonie Kreis, Anton Schiela},
		pdfkeywords = {Frobenius norm, inner product, trace estimation, neural network training, backpropagation, K-FAC preconditioner},
		pdfcreator = {Created using the Scoop Template Engine version 1.2.0.}
	}
}{
	\pdfinfo{
		/Title (Frobenius-Type Norms and Inner Products of Matrices and Linear Maps with Applications to Neural Network Training)
		/Author (Roland Herzog, Frederik Köhne, Leonie Kreis, Anton Schiela)
		/Subject ()
		/Keywords (Frobenius norm, inner product, trace estimation, neural network training, backpropagation, K-FAC preconditioner)
		/Creator (Created using the Scoop Template Engine version 1.2.0.)
	}
}
\makeatother

\title[Frobenius-Type Norms and Inner Products]{Frobenius-Type Norms and Inner Products of Matrices and Linear Maps with Applications to Neural Network Training}

\author{Roland Herzog\orcidlink{0000-0003-2164-6575}}
\address[R. Herzog]{Interdisciplinary Center for Scientific Computing, Heidelberg University, 69120 Heidelberg, Germany}
\email{roland.herzog@iwr.uni-heidelberg.de}
\urladdr{https://scoop.iwr.uni-heidelberg.de}

\author{Frederik Köhne\orcidlink{0009-0008-6185-9675}}
\address[F. Köhne]{Department of Mathematics, University of Bayreuth, 95440 Bayreuth, Germany}
\email{frederik.koehne@uni-bayreuth.de}
\urladdr{https://num.math.uni-bayreuth.de/en/team/frederik-koehne/}

\author{Leonie Kreis\orcidlink{0000-0003-4234-4867}}
\address[L. Kreis]{Interdisciplinary Center for Scientific Computing, Heidelberg University, 69120 Heidelberg, Germany}
\email{leonie.kreis@iwr.uni-heidelberg.de}
\urladdr{https://scoop.iwr.uni-heidelberg.de}

\author{Anton Schiela\orcidlink{0000-0002-6959-2951}}
\address[A. Schiela]{Department of Mathematics, University of Bayreuth, 95440 Bayreuth, Germany}
\email{anton.schiela@uni-bayreuth.de}
\urladdr{https://num.math.uni-bayreuth.de/en/team/anton-schiela/}

\thanks{This work was supported by DFG grants HE~6077/13--1 and SCHI~1379/8--1 within the Priority Program SPP~2298 (Mathematical Foundations of Deep Learning), which is gratefully acknowledged.}

\date{\today}

\dedicatory{}

\begin{document}

% Insert the abstract.
\begin{abstract}
The Frobenius norm is a frequent choice of norm for matrices.
In particular, the underlying Frobenius inner product is typically used to evaluate the gradient of an objective with respect to matrix variable, such as those occuring in the training of neural networks.
We provide a broader view on the Frobenius norm and inner product for linear maps or matrices, and establish their dependence on inner products in the domain and co-domain spaces.
This shows that the classical Frobenius norm is merely one special element of a family of more general Frobenius-type norms.
The significant extra freedom furnished by this realization can be used, among other things, to precondition neural network training.

\end{abstract}

% Insert the keywords.
\keywords{Frobenius norm, inner product, trace estimation, neural network training, backpropagation, K-FAC preconditioner}

% Insert the Mathematics Subject Classification.
\makeatletter
\ltx@ifpackageloaded{hyperref}{%
\subjclass[2010]{}
}{%
\subjclass[2010]{}
}
\makeatother

% Typeset the opening page.
\maketitle

% Insert the document body.
\section{Introduction}
\label{section:introduction}

The Frobenius norm is a frequently used tool to assess the magnitude of a matrix.
Given $S \in \K^{n \times m}$, where $\K$ is either the field of real or complex numbers, the Frobenius norm is defined as
\begin{equation}
	\label{eq:Frobenius-norm}
	\norm{S}_F
	\coloneqq
	\paren[Big](){\sum_{i=1}^n \sum_{j=1}^m \abs{s_{ij}}^2}^{1/2}
	=
	\paren[big](){\trace S^\hermitian S}^{1/2}
	.
\end{equation}
Equivalently, $\norm{S}_F$ can be conceived as the Euclidean norm of the vectorized form of~$S$, obtained by stacking, \eg, its columns.
The Frobenius norm arises from the Frobenius inner product,
\begin{equation}
	\label{eq:Frobenius-inner-product}
	\inner{S}{T}_F
	\coloneqq
	\sum_{i=1}^n \sum_{j=1}^m s_{ij} \, \conjugate t_{ij}
	=
	\trace T^\hermitian S
	=
	\trace S^\transp \conjugate{T}
	.
\end{equation}

In this short paper, we provide a broader view on the notions of Frobenius inner product and norm.
Notably, we establish that these notions extend to linear maps between finite-dimensional vector spaces, and that they naturally depend on the choice of inner products in the domain and co-domain spaces $\cV$ and $\cW$.
When these inner products are represented by Hermitian positive definite matrices~$V$ and $W$, respectively, we obtain the family of Frobenius-type inner products
\begin{equation}
	\label{eq:Frobenius-type-inner-product-in-terms-of-matrices}
	\trace \conjugate{V}^{-1} \! S^\transp W \conjugate{T}
\end{equation}
of $S$ and $T$ as a generalization of \eqref{eq:Frobenius-inner-product}.

The significant extra freedom coming from the choice of inner products in the domain and co-domain spaces can be leveraged in many ways, for instance, in optimization problems involving matrix variables, such as those underlying the training of artificial neural networks.
While typically the standard Frobenius inner product \eqref{eq:Frobenius-inner-product} is used to define the gradient of the objective \wrt these variables, the more general Frobenius-type inner product \eqref{eq:Frobenius-type-inner-product-in-terms-of-matrices} admits significant extra flexibility.
When this is used to construct data dependent inner product matrices~$V$ and $W$, the resulting preconditioned gradient descent can lead to substantial improvements in convergence.
The K-FAC preconditioner by \cite{MartensGrosse:2015:1} can be interpreted in this way.

A second finding of this paper is that the Frobenius-type norms have interpretations similar to an operator norm of a linear map.
While the usual operator norm is defined as the maximal norm response of the map \wrt inputs in the unit sphere, the Frobenius norm turns out to measure the average norm response.

This paper is outlined as follows.
\Cref{section:preliminaries} reviews the concept of traces for endomorphisms of finite dimensional vector spaces.
It also introduces the metric trace for sesquilinear forms, equivalent to conjugate linear maps of the vector space into its dual.
As the name suggests, the metric trace depends on the inner product.
The Frobenius-type inner product and associated norm follows as a conclusion of these considerations.
In \cref{section:metric-trace-and-Frobenius-type-norm-via-averaging} we interpret the metric trace and the Frobenius-type norm in the sense similar to the operator norm of a linear map, but with maximal response replaced by average response.
Finally, in \cref{section:preconditioned-gradient-methods-for-neural-network-training}, we provide a brief overview into how the concept of Frobenius-type inner products leads to the notion of preconditioned gradient directions for objective functions \wrt matrix variables.
We conclude with an outlook into how this concept can be used in (stochastic) gradient descent algorithms for neural network training, and interpret the K-FAC preconditioner in this light.

In \cref{section:preliminaries,section:metric-trace-and-Frobenius-type-norm-via-averaging} we use notation applicable when $\K = \C$, which is more general than $\K = \R$.
The reader may safely particularize all results to the case of real vector spaces, maps and matrices by ignoring conjugation (overbars) and by replacing \enquote{Hermitian} by \enquote{symmetric}, \enquote{sesquilinear} by \enquote{bilinear}, \enquote{conjugate linear} and \enquote{antilinear} by \enquote{linear}, and $\cdot^\hermitian$ by the transpose~$\cdot^\transp$.

\section{Preliminaries}
\label{section:preliminaries}

In this section we recall some known facts from linear algebra, concerning linear maps and their representation by matrices.
Throughout, $\K$ denotes the field of real or complex numbers.
Symbols such as $\cV$ and $\cW$ denote finite-dimensional vector spaces over~$\K$.
The space of linear maps from $\cV$ to $\cW$ is referred to as $\cL(\cV,\cW)$.
We denote linear maps between such spaces by $\cS$ and $\cT$ \etc
The space $\cL(\cV,\K)$ is denoted by $\cV^*$ and termed the \emph{dual space} of $\cV$.
We denote the dual pairing of $\ell \in \cV^*$ and $v \in \cV$ by $\dual{\ell}{v}$ or $\dual{v}{\ell}$.
The adjoint of a linear map $\cS \in \cL(\cV,\cW)$ is the linear map $\cS^* \in \cL(\cW^*,\cV^*)$ defined by $\dual{\cS v}{\ell} = \dual{v}{\cS^* \ell}$ for all $v \in \cV$ and $\ell \in \cW^*$.

Given a basis $\{v_j\}_{j = 1, \dots, n}$ of $\cV$ and a basis $\{w_i\}_{i = 1, \dots, m}$ of $\cW$, a linear map $\cS \in \cL(\cV,\cW)$ is represented by the matrix $S = (S_{ij})$ with entries $S_{ij} \in \K$ satisfying
\begin{equation}
	\label{eq:linear-map:matrix-representation}
	\cS v_j
	=
	\sum_{i=1}^m S_{ij} \, w_i
	.
\end{equation}
In the case of endomorphisms, \ie, elements of $\cL(\cV,\cV)$, we will consistently use the same basis for both copies of~$\cV$.
Specifically, $\id_\cV$ denotes the identity map in $\cL(\cV,\cV)$.

An antilinear, or conjugate linear map $\cA \colon \cV \to \cV^*$ is a map satisfying
\begin{equation*}
	\cA \, (\alpha_1 v_1 + \alpha_2 v_2)
	=
	\conjugate{\alpha_1} \cA v_1
	+
	\conjugate{\alpha_2} \cA v_2
\end{equation*}
for all $\alpha_1, \alpha_2 \in \K$ and $v_1, v_2 \in \cV$.
We denote the vector space of conjugate linear maps from $\cV$ into $\cW$ by $\conjugate{\cL}(\cV,\cW)$ and observe that the composition of two conjugate linear maps is linear. 
Be aware that \emph{conjugate} linear maps in $\cA \in \conjugate{\cL}(\cV,\cV^*)$ also have matrix representations $A = (A_{ij})$ with entries $A_{ij} \in \K$ satisfying
\begin{equation}
	\label{eq:conjugate-linear-map:matrix-representation}
	A_{ij}
	=
	\dual{v_i}{\cA \, v_j}
	.
\end{equation}
This representation works as follows.
When $v = \sum\limits_{j=1}^n \alpha_j \, v_j$ and $\widehat v = \sum\limits_{i=1}^n \beta_i \, v_i$, we understand
\begin{equation*}
	\begin{aligned}
		\dual{\widehat v}{\cA \, v}
		&
		=
		\dual[Big]{\sum_{i=1}^n \beta_i \, v_i}{\cA \, \sum_{j=1}^n \alpha_j \, v_j}
		=
		\dual[Big]{\sum_{i=1}^n \beta_i \, v_i}{\sum_{j=1}^n \conjugate{\alpha_j} \, \cA \, v_j}
		\\
		&
		=
		\sum_{i,j=1}^n \beta_i \, \conjugate{\alpha_j} \dual{v_i}{\cA \, v_j}
		=
		\sum_{i,j=1}^n \beta_i \, \conjugate{\alpha_j} A_{ij}
		=
		\beta^\transp A \, \conjugate{\alpha}
		=
		\alpha^\hermitian A^\transp \, \beta
		.
	\end{aligned}
\end{equation*}
More generally, for $\cA \in \conjugate{\cL}(\cV,\cW)$, $v = \sum\limits_{j=1}^n \alpha_j \, v_j$ and $\cA \, v = \sum\limits_{i=1}^m \beta_i \, w_i$, $\cA$ is represented by the matrix~$A$ satisfying
\begin{equation}
	\label{eq:conjugate-linear-map:matrix-representation:2}
	\beta
	=
	A \, \conjugate{\alpha}
	.
\end{equation}

The representations \eqref{eq:linear-map:matrix-representation}, \eqref{eq:conjugate-linear-map:matrix-representation} and \eqref{eq:conjugate-linear-map:matrix-representation:2} of linear and conjugate linear maps are compatible with composition in the following sense.

When $\cS \in \cL(\cV,\cW)$ is represented by $S$ as in \eqref{eq:linear-map:matrix-representation} and $\cA \in \conjugate{\cL}(\cW,\cX)$ is represented by $A$ as in \eqref{eq:conjugate-linear-map:matrix-representation:2}, then $\cA \, \cS \in \conjugate{\cL}(\cV,\cX)$ is represented by $A \, \conjugate{S}$.
When $\cB \in \conjugate{\cL}(\cV,\cW)$ is represented by $B$ as in \eqref{eq:conjugate-linear-map:matrix-representation:2}, then $\cA \, \cB \in \cL(\cV,\cX)$ is represented by $A \, \conjugate{B}$.
As a rule, composition of any map, linear or conjugate linear, with a conjugate linear map on the left is represed by the product of the two respective matrices, with the factor on the right conjugted.

For instance, in \eqref{eq:Frobenius-type-inner-product-in-terms-of-matrices}, $S$ and $T$ represent elements $\cS$ and $\cT$ of $\cL(\cV,\cW)$, and $S^\transp$ represents the adjoint of $\cS \in \cL(\cW^*,\cV^*)$ \wrt the dual bases.
Moreover, $V$ and $W$ represent the Riesz maps of $\cV$ and $\cW$, which are elements of $\conjugate{\cL}(\cV,\cV^*)$ and $\conjugate{\cL}(\cW,\cW^*)$, respectively. 

When $\K = \R$, there is no distinction between linear and conjugate linear maps, \ie, between $\cL(\cV,\cV^*)$ and $\conjugate{\cL}(\cV,\cV^*)$.

\subsection{Traces of Endomorphisms}
\label{subsection:traces-of-endomorphisms}

Classically, on every finite dimensional vector space~$\cV$, a \emph{trace} is defined as a linear map from the vector space of endomorphisms $\cL(\cV,\cV)$ of $\cV$ into the corresponding field, \ie,
\begin{equation*}
	\trace 
	\colon
	\cL(\cV,\cV) 
	\to 
	\K
	;
\end{equation*}
see for instance \cite[Chapter~II, §10.11]{Bourbaki:1998:1} or \cite[Section~8D]{Axler:2023:1}.
Besides linearity, the trace is completely characterized by the commutativity property 
\begin{equation}
	\label{eq:trace_commutes}
	\trace \cS \, \cT 
	= 
	\trace \cT \cS
\end{equation}
for all $\cS, \cT \in \cL(\cV,\cV)$, together with the normalization condition
\begin{equation*}
	\trace \id_\cV 
	= 
	\dim \cV
	.
\end{equation*}
In fact, \eqref{eq:trace_commutes} holds more generally, for all $\cS \in \cL(\cV,\cW)$ and $\cT \in \cL(\cW,\cV)$.
It follows immediately that $\trace$ is invariant under cyclic permutations of three (or more) linear maps, \ie, we have
\begin{equation}
	\label{eq:trace-is-cyclic}
	\trace \cT \cS \, \cU
	= 
	\trace \cS \, \cU \cT
	= 
	\trace \cU \, \cT \cS
\end{equation}
for all $\cS \in \cL(\cV,\cW)$, $\cT \in \cL(\cW,\cX)$, and $\cU \in \cL(\cX,\cV)$.

The trace of an endomorphism $\cE \in \cL(\cV,\cV)$ has the well-known explicit representation
\begin{equation}
	\label{eq:trace_explicit}
	\trace \cE
	= 
	\trace E
	\coloneqq
	\sum_{j=1}^n E_{jj}
\end{equation}
in terms of the matrix~$E$ representing~$\cE$ \wrt an arbitrary basis of $\cV$.
In fact, a change of basis is affected by replacing $E$ by its similarity transform $F^{-1} E F$, and 
\begin{equation*}
	\trace F^{-1} E F 
	= 
	\trace E F F^{-1} 
	= 
	\trace E
\end{equation*}
shows that $\trace \cE$ is indeed a property of the endomorphism, independent of the choice of basis in $\cV$ and thus independent from the particular representation of~$\cE$ as a matrix~$E$. 
In addition, we remark that $\trace \cE$ coincides with the sum of the eigenvalues of the endomorphism~$\cE$ as well with the sum of the eigenvalues of any of its representing matrices~$E$.

We noted above that the composition of two conjugate linear maps is linear, so we can study the analog of the commutativity property \eqref{eq:trace_commutes} for conjugate linear maps:
\begin{lemma}
	\label{lemma:trace-is-anti-cyclic-for-conjugate-linear-maps}
	Suppose that $\cA \in \conjugate{\cL}(\cV,\cW)$, $\cB \in \conjugate{\cL}(\cV,\cW)$ are conjugate linear maps.
	Then
	\begin{equation*}
		\trace \cA \, \cB 
		= 
		\conjugate{\trace \cB \, \cA}
		.
	\end{equation*}
\end{lemma}
\begin{proof}
	Suppose that $A$ and $B$ are the representations of $\cA$ and $\cB$, respectively, \wrt some bases in $\cV$ and $\cW$, respectively; see \eqref{eq:conjugate-linear-map:matrix-representation:2}.
	Then $\cA \, \cB$ is represented by $A \, \conjugate{B} = \conjugate{\conjugate{A} \, B}$ and $\cB \cA$ is represented by $B \, \conjugate{A}$.
	Therefore, 
	\begin{equation*}
		\trace \cA \, \cB 
		=
		\trace A \, \conjugate{B}
		=
		\trace \conjugate{\conjugate{A} \, B}
		=
		\conjugate{\trace \conjugate{A} \, B}
		=
		\conjugate{\trace B \, \conjugate{A}}
		=
		\conjugate{\trace \cB \cA}
		.
		\qedhere
	\end{equation*}
\end{proof}

\subsection{Metric Trace of a Sesquilinear Form}
\label{subsectoin:metric-trace-of-a-sesquilinear-form}

In the previous subsection we recalled that the trace is a property of endomorphisms $\cL(\cV,\cV)$ on the vector space~$\cV$ that is oblivious of any additional structure such as an inner product which may be present in~$\cV$.
When an inner product $\inner{\cdot}{\cdot}_\cV$ is present in~$\cV$, we can devise a notion of trace for conjugate linear maps $\cA \in \conjugate{\cL}(\cV,\cV^*)$ from $\cV$ into its dual, or equivalently, for sesquilinear forms.
To emphasize that the trace for sesquilinear forms depends on the inner product, we refer to it as the \emph{metric trace}.

A sesquilinear form over $\cV$ is a map $a \colon \cV \times \cV \to \K$ that is linear in the first and antilinear in the second argument, \ie,
\begin{subequations}
	\label{eq:sesquilinear-form}
	\begin{align}
		a(\alpha_1 v_1 + \alpha_2 v_2, v)
		&
		=
		\alpha_1 \, a(v_1,v)
		+
		\alpha_2 \, a(v_2,v)
		\\
		a(v,\alpha_1 v_1 + \alpha_2 v_2)
		&
		=
		\conjugate{\alpha_1} \, a(v,v_1)
		+
		\conjugate{\alpha_2} \, a(v,v_2)
		&
	\end{align}
\end{subequations}
for all $\alpha_1, \alpha_2 \in \K$ and $v_1, v_2, v \in \cV$.
Notice that, due to the polarization identity, a sesquilinear form is completely determined by its quadratic form $a(v,v)$, \ie, by its values on the diagonal.

Sesquilinear forms are in a bijective relationship with conjugate linear maps in $\conjugate{\cL}(\cV,\cV^*)$ via
\begin{equation}
	\label{eq:sesquilinear-form-and-associated-conjugate-linear-map}
	a(v_1,v_2)
	=
	\dual{\cA \, v_2}{v_1}
	.
\end{equation}

Suppose now that $\inner{\cdot}{\cdot}_\cV$ is an inner product on $\cV$, \ie, a sesquilinear form which is also Hermitian and positive definite.
That is, we have $\inner{v_1}{v_2}_\cV = \conjugate{\inner{v_2}{v_1}_\cV}$ and $\inner{v}{v}_\cV > 0$ for all $v \neq 0$.
The inner product induces the Riesz map $\cR_\cV \in \conjugate{\cL}(\cV,\cV^*)$, defined by
\begin{equation*}
	\dual{\cR_\cV v}{w}
	=
	\inner{w}{v}_\cV
\end{equation*}
for all $v, w \in \cV$.
Observe that $\cR_\cV$ is not linear but conjugate linear, \ie,
\begin{equation*}
	\cR_\cV(\alpha \, v)
	=
	\conjugate{\alpha} \, \cR_\cV v
\end{equation*}
holds for $v \in \cV$ and $\alpha \in \K$.
A particular case of inner product is the standard inner product $\inner{v_1}{v_2}_{\K^n} = v_2^\hermitian v_1 = v_1^\transp \conjugate{v_2}$ of $\cV = \K^n$, represented by the identity matrix.

We can now introduce the metric trace for a sesquilinear form $a \colon \cV \times \cV \to \K$ and, simultaneously, the metric trace for the conjugate linear map $\cA \in \conjugate{\cL}(\cV,\cV^*)$ associated to $a$ via \eqref{eq:sesquilinear-form-and-associated-conjugate-linear-map}.
We define the metric trace for $a$ and $\cA$ as
\begin{equation}
	\label{eq:metric-trace}
	\trace_\cV a
	\coloneqq
	\trace_\cV \cA
	\coloneqq
	\conjugate{\trace \cR_\cV^{-1} \cA}
	.
\end{equation}
Notice that $\cR_\cV^{-1} \cA$ is an endomorphism on $\cV$. 
Due to \cref{lemma:trace-is-anti-cyclic-for-conjugate-linear-maps}, we also have $\trace_\cV a = \trace{\cA \, \cR_\cV^{-1}}$.

The definition of metric trace has the following properties:
\begin{lemma}
	\label{lemma:metric-trace}
	Suppose that $\inner{\cdot}{\cdot}_\cV$ is an inner product on $\cV$.
	The metric trace \eqref{eq:metric-trace} \wrt this inner product has the following properties:
	\begin{enumerate}
		\item \label[property]{item:metric-trace:1}
			$\trace_\cV$ is linear, \ie,
			\begin{equation}
				\label{eq:metric-trace:linearity}
				\trace_\cV (\alpha_1 \, a_1 + \alpha_2 \, a_2)
				=
				\alpha_1 \trace_\cV a_1 + \alpha_2 \trace_\cV a_2
			\end{equation}
			for all $\alpha_1, \alpha_2 \in \K$ and sesquilinear forms~$a_1, a_2 \colon \cV \times \cV \to \K$.

		\item \label[property]{item:metric-trace:2}
			When $\{q_i\}_{i = 1, \dots, n}$ is an orthonormal basis of $\cV$, then
			\begin{equation}
				\label{eq:traceSum}
				\trace_\cV a 
				= 
				\sum_{i=1}^n a(q_i,q_i)
				.
			\end{equation}
	\end{enumerate}
\end{lemma}
\begin{proof}
	The equality
	\begin{align*}
		\MoveEqLeft
		\trace_\cV (\alpha_1 \, a_1 + \alpha_2 \, a_2)
		\\
		&
		= 
		\conjugate{\trace(\cR_\cV^{-1} (\alpha_1 \cA_1 + \alpha_2 \cA_2))}
		=
		\conjugate{\trace(\conjugate{\alpha_1} \, \cR_\cV^{-1} \cA_1 + \conjugate{\alpha_2} \, \cR_\cV^{-1} \cA_2)}
		\\
		&
		=
		\alpha_1 \trace a_1
		+
		\alpha_2 \trace a_2
	\end{align*}
	proves \cref{item:metric-trace:1}.

	The linear mapping $\cQ \colon \K^n \to \cV$, defined by $\cQ e_i = q_i$, is an isometry due to 
	\begin{equation*}
		\inner{\cQ \, v_1}{\cQ \, v_2}_\cV
		=
		\inner{v_1}{v_2}_{\K^n}
		=
		\conjugate{v_1^\hermitian v_2}
		.
	\end{equation*}
	We use this to compute
	\begin{align*}
		\sum_{i=1}^n a(q_i,q_i)
		&
		=
		\sum_{i=1}^n \dual{\cA q_i}{q_i}
		=
		\sum_{i=1}^n \inner{q_i}{\cR_\cV^{-1} \cA q_i}_\cV
		=
		\sum_{i=1}^n \inner{\cQ e_i}{\cQ \cQ^{-1} \cR_\cV^{-1} \cA \cQ e_i}_\cV
		\\
		&
		=
		\sum_{i=1}^n \conjugate{e_i^\hermitian \cQ^{-1} \cR_\cV^{-1} \cA \cQ e_i}
		=
		\conjugate{\trace \cQ^{-1} \cR_\cV^{-1} \cA \cQ}
		=
		\conjugate{\trace \cR_\cV^{-1} \cA}
		,
	\end{align*}
	which is \cref{item:metric-trace:2}.
\end{proof}

Note that if $\cE \in \cL(\cV,\cV)$ is an endomorphism and $\inner{\cdot}{\cdot}_\cV$ is an inner product on~$\cV$, then for any orthonormal basis $\{q_i\}_{i = 1, \dots, n}$, we have
\begin{equation*}
	\trace \cE
	= 
	\conjugate{\trace_\cV \cR_\cV \cE}
	=
	\sum_{i=1}^n \conjugate{\inner{q_i}{\cE q_i}_\cV}
	=
	\sum_{i=1}^n \inner{\cE q_i}{q_i}_\cV
	.
\end{equation*}
On $\K^n$, equipped with the standard inner product, we retain the original representation of the trace \eqref{eq:trace_explicit} by inserting the basis of unit vectors:
\begin{equation*}
	\trace
	\cE
	=
	\sum_{i=1}^n \inner{e_i}{\cE e_i}_{\K^n}
	=
	\sum_{i=1}^n e_i^\transp E \, \conjugate{e_i}
	= 
	\sum_{i=1}^n e_i^\transp E \, e_i
	= 
	\sum_{i=1}^n E_{ii}
	.
\end{equation*}

\subsection{Frobenius-Type Inner Product and Norm}
\label{subsection:frobenius-type-inner-product-and-norm}

In this section we generalize the notion of Frobenius inner product and Frobenius norm from matrices to linear maps $\cS, \cT \in \cL(\cV,\cW)$.
This notion depends on inner products $\inner{\cdot}{\cdot}_\cV$ and $\inner{\cdot}{\cdot}_\cW$ in both spaces.

To begin with, we define a sesquilinear map~$a \colon \cV \times \cV \to \K$ using the data $\cS, \cT$ and $\inner{\cdot}{\cdot}_\cW$ via
\begin{equation*}
	a(v_1,v_2)
	\coloneqq
	\inner{\cS v_1}{\cT v_2}_\cW
	=
	\dual{v_1}{\cS^* \cR_\cW \cT v_2}
	.
\end{equation*}
The conjugate linear map associated with~$a$ is $\cS^* \cR_\cW \cT \in \conjugate{\cL}(\cV,\cV^*)$.
Now the Frobenius-type inner product of $\cS$ and $\cT$ induced by the inner products $\inner{\cdot}{\cdot}_\cV$ and $\inner{\cdot}{\cdot}_\cW$ is defined as
\begin{equation}
	\label{eq:Frobenius-inner-product-for-linear-maps}
	\inner{\cS}{\cT}_{\cV \to \cW}
	\coloneqq 
	\trace_\cV a
	=
	\trace_\cV \cS^* \cR_\cW \cT
	=
	\conjugate{\trace \cR_\cV^{-1} \cS^* \cR_\cW \cT}
	.
\end{equation}
For any orthonormal basis $\{q_i\}_{i = 1, \ldots, n}$ of $\cV$, \cref{lemma:metric-trace} yields
\begin{equation}
	\label{eq:Frobenius-inner-product-for-linear-maps:orthonormal}
	\inner{\cS}{\cT}_{\cV \to \cW}
	=
	\sum_{i=1}^n \inner{\cS q_i}{\cT q_i}_\cW
	.
\end{equation}
This shows that \eqref{eq:Frobenius-inner-product-for-linear-maps} is indeed an inner product, \ie, a sesquilinear and positive definite form on $\cL(\cV,\cW)$.
In addition, we can infer using \cref{lemma:trace-is-anti-cyclic-for-conjugate-linear-maps}, that
\begin{equation}
	\label{eq:Frobenius-inner-product-of-adjoints}
	\inner{\cS^*}{\cT^*}_{\cW^* \to \cV^*}
	=
	\inner{\cS}{\cT}_{\cV \to \cW}
\end{equation}
holds.

The Frobenius-type norm of $\cT$ induced by this inner product can now be defined via
\begin{equation}
	\label{eq:Frobenius-norm-for-linear-maps}
	\norm{\cT}_{\cV \to \cW}
	\coloneqq
	\inner{\cT}{\cT}_{\cV \to \cW}^{1/2}
	=
	\paren[Big](){\sum_{i=1}^n \inner{\cT q_i}{\cT q_i}_\cW}^{1/2}
	.
\end{equation}
\Cref{eq:Frobenius-inner-product-of-adjoints} shows that $\norm{\cT^*}_{\cW^* \to \cV^*} = \norm{\cT}_{\cV \to \cW}$ holds as well.

As a special case, we recover from \eqref{eq:Frobenius-norm-for-linear-maps} the standard Frobenius inner product \eqref{eq:Frobenius-inner-product} and associated norm \eqref{eq:Frobenius-norm}, using the choice $\cV = \K^n$ and $\cW = \K^m$, both endowed with the standard inner products:
\begin{equation*}
	\inner{\cS}{\cT}_{\K^n \to \K^m}
	=
	\sum_{i=1}^n \inner{\cS e_i}{\cT e_i}_{\K^m}
	=
	\sum_{i=1}^n \inner{S e_i}{T e_i}_{\K^m}
	=
	\sum_{i=1}^n e_i^\transp S^\transp \conjugate{T e_i}
	= 
	\trace S^\transp \conjugate{T}
	.
\end{equation*}
More generally, when $\cV = \K^n$ and $\cW = \K^m$ are endowed with inner products represented by matrices~$V$ and $W$ \wrt the standard bases, \ie, $\inner{v_1}{v_2}_\cV = v_1^\transp V \conjugate{v_2}$ and accordingly for~$W$, we obtain \eqref{eq:Frobenius-type-inner-product-in-terms-of-matrices} from the definition \eqref{eq:Frobenius-inner-product-for-linear-maps}.
Indeed, 
\begin{equation*}
	\conjugate{\trace \cR_\cV^{-1} \cS^* \cR_\cW \cT}
	=
	\conjugate{\trace V^{-1} \conjugate{S^\transp W \conjugate{T}}}
	=
	\conjugate{\trace V^{-1} \conjugate{S}^\transp \conjugate{W} T}
	=
	\trace \conjugate{V}^{-1} S^\transp W \conjugate{T}
	.
\end{equation*}

\section{Interpretation of the Metric Trace and Frobenius-Type Norm via Averaging}
\label{section:metric-trace-and-Frobenius-type-norm-via-averaging}

In this section we provide an interpretation of the metric trace of a sesquilinear form $a \colon \cV \times \cV \to \K$ in the sense of an average response of its quadratic form $v \mapsto a(v,v)$.
Since Frobenius-type inner products \eqref{eq:Frobenius-inner-product-for-linear-maps} and associated norms are defined in terms of metric traces, we obtain a similar interpretation for them as well.

Suppose now that~$a$ is a sesquilinear form over $\cV$ and $\cA \in \conjugate{\cL}(\cV,\cV^*)$ is the associated conjugate linear map; see \eqref{eq:sesquilinear-form-and-associated-conjugate-linear-map}.
Any $v \in \cV$ can be viewed as the canonical linear map $v \in \cL(\K,\cV)$ via $\alpha \mapsto \alpha \, v$.
On the other hand, $v$ generates the conjugate linear map $v^\bidual \in \conjugate{\cL}(\cV^*,\K)$ via $\ell \mapsto \conjugate{\dual{v}{\ell}}$.
We can therefore write
\begin{equation}
	\label{eq:rewrite-sesquilinear-form-using-the-trace}
	a(v,v)
	=
	\dual{\cA v}{v}
	=
	\conjugate{\dual{v^\bidual}{\cA v}}
	=
	\conjugate{v^\bidual \cA v}
	=
	\conjugate{\trace (v^\bidual \cA v)}
	=
	\conjugate{\trace (v \, v^\bidual \cA)}
\end{equation}
for any $v \in \cV$.
Note that $v^\bidual \cA v$, which is just an alternative notation for $\dual{v^\bidual}{\cA v}$, belongs to $\cL(\K,\K) \cong \K$ and thus it is its own trace.

Now suppose that $v$ is a $\cV$-valued random variable with expected value $\E(v) = 0$ and covariance $\cC \coloneqq \E(v \, v^\bidual) \in \conjugate{\cL}(\cV^*,\cV)$.
The sesquilinear form associated with the covariance is $c(\ell_1,\ell_2) = \E(\dual{v \, v^\bidual \ell_2}{\ell_1}) = \E(\conjugate{\dual{v}{\ell_2}} \dual{v}{\ell_1})$ and it is thus easily seen to be Hermitian and positive semidefinite.

Using these considerations, we can now prove the following
\begin{theorem}
	\label{theorem:metric-trace-of-sesquilinear-form-is-expected-value-of-quadratic-form}
	Suppose that $v$ is a $\cV$-valued random variable with expected value $\E(v) = 0$ and covariance $\cC \coloneqq \E(v \, v^\bidual)$.
	Then for any sesquilinear form~$a(\cdot,\cdot)$ over~$\cV$ and associated conjugate linear map~$\cA$, we have $\cC \cA \in \cL(\cV,\cV)$ and
	\begin{equation}
		\label{eq:covariance-preconditioned-operator-is-expected-value-of-quadratic-form}
		\conjugate{\trace (\cC \cA)}
		=
		\E(a(v,v))
		.
	\end{equation}
	Moreover, if $\cC$ is positive definite and thus induces the inner product $\inner{v_1}{v_2}_{\cC^{-1}} \coloneqq \dual{\cC^{-1} v_1}{v_2}$ on~$\cV$ with Riesz map $\cR_\cV = \cC^{-1}$, we have
	\begin{equation}
		\label{eq:metric-trace-of-sesquilinear-form-is-expected-value-of-quadratic-form}
		\trace_{\cC^{-1}} a
		=
		\E(a(v,v))
		.
	\end{equation}
\end{theorem}
\begin{proof}
	Taking the expected value in \eqref{eq:rewrite-sesquilinear-form-using-the-trace} shows \eqref{eq:covariance-preconditioned-operator-is-expected-value-of-quadratic-form}:
	\begin{equation*}
		\E(a(v,v))
		=
		\E(\conjugate{\trace (v \, v^\bidual \cA)})
		=
		\conjugate{\trace (\E(v \, v^\bidual) \cA)}
		=
		\conjugate{\trace (\cC \cA)}
		.
	\end{equation*}
	Under the assumption that $\cC$ is positive definite, \eqref{eq:metric-trace-of-sesquilinear-form-is-expected-value-of-quadratic-form} is a consequence of the definition \eqref{eq:metric-trace} of metric trace.
\end{proof}

We point out that \cref{theorem:metric-trace-of-sesquilinear-form-is-expected-value-of-quadratic-form} is the basis of the well known trace estimation technique \cite{Hutchinson:1989:1} in the special case that $\cC^{-1}$ represents the standard inner product in~$\K^n$.
As advertised in the beginning of the section, \eqref{eq:metric-trace-of-sesquilinear-form-is-expected-value-of-quadratic-form} provides an interpretation of the metric trace of the sesquilinear form~$a(\cdot,\cdot)$ as the expected, or average response of its quadratic form $v \mapsto a(v,v)$, provided that the inner product used in the definition of the metric trace is the inverse covariance of the probability measure governing the distribution of $v \in \cV$.
This inverse covariance is often referred to as the precision operator in statistics.

The definition of metric trace \eqref{eq:metric-trace} of a conjugate linear map $\cA \in \conjugate{\cL}(\cV,\cV^*)$ and thus the Frobenius-type inner product of elements of $\cL(\cV,\cW)$ involves the inverse Riesz map of the space~$\cV$.
The following theorem provides a way to evaluate this.

\begin{theorem}
	\label{theorem:covariance-as-surface-integral}
	Consider $\cV$ with an inner product $\inner{\cdot}{\cdot}_\cV$ and $\dim \cV = n$. 
	Assume that $\mu$ is a measure on $\cV$ that is invariant against linear isometric isomorphisms $\cT \in \cL(\cV,\cV$), \ie, $\mu(M) = \mu_\cT(M) \coloneqq \mu(\cT^{-1}(M))$ for all measurable $M \subset \cV$.
	Then the corresponding Riesz isomorphism~$\cR_{\cV}$ satisfies
	\begin{equation}
		\label{eq:covariance-as-general-integral}
		\cR_{\cV}^{-1}
		=
		\frac{n}{\abs{\mu}} \int \frac{v \, v^\bidual}{\norm{v}_\cV^2} \d \mu
		,
	\end{equation}
	where $\abs{\mu} = \mu(\cV)$ denotes the $\mu$-volume of $\cV$, assumed finite. 
\end{theorem}
\begin{proof}
	Consider the conjugate linear mapping $\cA \in \conjugate{\cL}(\cV^*,\cV)$ defined by $\int v \, v^\bidual/\norm{v}_\cV^2 \d \mu$.
	For any linear mapping $\cQ \in \cL(\cV,\cV)$ with adjoint $\cQ^* \in \cL(\cV^*,\cV^*)$ we first observe
	\begin{equation*}
		(\cQ \, v)^\bidual \ell
		=
		\ell(\cQ \, v)
		=
		(\cQ^*\ell)(v)
		=
		v^\bidual \cQ^*\ell 
		,
		\quad
		\text{\ie}
		,
		\quad
		(\cQ \, v)^\bidual
		=
		v^\bidual \cQ^*
		.
	\end{equation*}
	Assuming that $\cQ$ is an isometric isomorphism and using the linearity of the integral, we can write
	\begin{equation*}
		\cQ \cA \cQ^*
		=
		\cQ \int \frac{v \, v^\bidual}{\norm{v}_\cV^2} \d \mu\, \cQ^*
		=
		\int \frac{\cQ \, v \, v^\bidual \cQ^*}{\norm{v}_\cV^2} \d \mu  
		=
		\int \frac{(\cQ \, v) (\cQ \, v)^\bidual} {\norm{v}_\cV^2} \d \mu
		.
	\end{equation*}
	Using the linear substitution $w = \cQ \, v$ we continue, exploiting the transformation formula for pushforward measures and the assumed invariance $\mu = \mu_{\cQ}$, we obtain
	\begin{equation*}
		\cQ \cA \cQ^*
		=
		\int \frac{w \, w^\bidual} {\norm{w}_\cV^2} \d \mu_{\cQ} 
		=
		\int \frac{w \, w^\bidual} {\norm{w}_\cV^2} \d \mu. 
	\end{equation*}
	This finally yields: 
	\begin{equation*}
		\cQ \cA \cQ^*
		=
		\cA
		.
	\end{equation*}
	Using the isometry property $\cQ^* \cR_\cV \cQ = \cR_\cV$ and defining $\cM \coloneqq \cA \, \cR_\cV \in \cL(\cV,\cV)$, we obtain that $\cM$ commutes with any isometry:
	\begin{equation*}
		\cQ \cM 
		=
		\cM \cQ
		.
	\end{equation*}
	By \cref{lemma:commuting-with-all-isometries} below, we conclude $\cM = \alpha \, \id_\cV$. 
	Finally, to determine $\alpha$, we can use the following straightforward computation: 
	\begin{align*}
		\alpha \, n 
		&
		=
		\trace \cM
		=
		\trace \cA \, \cR_\cV 
		= 
		\trace \paren[auto](){\int \frac{v \, v^\bidual}{\norm{v}_\cV^2} \d \mu \, \cR_\cV}
		\\
		&
		=
		\int \trace \paren[auto](){\frac{v \, v^\bidual}{\norm{v}_\cV^2} \cR_\cV} \d \mu
		= 
		\int \frac{v^\bidual  \cR_\cV v}{\norm{v}_\cV^2} \d \mu
		=
		\int 1 \d \mu
		=
		\abs{\mu}
		.
	\end{align*}
	This proves $\alpha = \frac{\abs{\mu}}{n}$ and thus $\cR_\cV^{-1} = \frac{n}{\abs{\mu}} \cA$.
\end{proof}

In the following, we denote the unit sphere and ball in $\cV$ \wrt an inner product $\inner{\cdot}{\cdot}_\cV$ by
\begin{equation}
	\label{eq:inverse-covariance-unit-sphere-and-ball}
	\sphere
	\coloneqq
	\setDef[big]{v \in \cV}{\inner{v}{v}_\cV = 1}
	\quad
	\text{and}
	\quad
	\ball
	\coloneqq
	\setDef[big]{v \in \cV}{\inner{v}{v}_\cV \le 1}
	.
\end{equation}
We denote the surface area of $\sphere$ by $\abs{\sphere}$ and the volume of $\ball$ by $\abs{\ball}$, both measured \wrt $\inner{\cdot}{\cdot}_\cV$.

\begin{corollary}\label{cor:various-measures}
	We have the identities:
	\begin{align*}
		\cR_{\cV}^{-1}
		&
		=
		\frac{n}{\abs{\sphere}} \int_{\sphere} v \, v^\bidual \d S
		\\
		&
		= 
		\frac{n}{\abs{\ball}} \int_{\ball} \frac{v \, v^\bidual}{\norm{v}_\cV^2} \d V
		\\
		&
		= 
		\frac{1}{(2\pi)^{n/2}} \int_\cV \frac{v \, v^\bidual}{\norm{v}_\cV^2} \, \e^{- \frac{1}{2} \norm{v}_\cV^2} 
		\d V
		.
	\end{align*}
	Here $\d V$ and $\d S$ are volume and surface measures, induced by the given inner product $\inner{\cdot}{\cdot}_{\cV}$.
\end{corollary}
\begin{proof}
	We apply the previous theorem to the surface measure of the sphere, induced by the inner product, the volume measure on the unit ball, also induced by the inner product, and the measure, induced by the probability density function of the normal distribution with covariance matrix $\cR_\cV^{-1}$. 
	All these measures are invariant against isometric isomorphisms: their supports are invariant against isometries. 
	Moreover, by definition $\d V(Q_n) = 1$ for any $n$-cube with edges of unit length in $\cV$, and similarly, $\d S(Q_{n-1}) = 1$ for any $Q_{n-1}$ cube with edges of unit length in $\cV$. 
	By construction, these objects are invariant against isometries. 
\end{proof}

\begin{lemma}
	\label{lemma:commuting-with-all-isometries}
	Consider $\cV$ with an inner product $\inner{\cdot}{\cdot}_\cV$.
	A linear map $\cM \in \cL(\cV,\cV)$ commutes with all isometries $\cQ \in \cL(\cV,\cV)$, \ie, $\cM \cQ = \cQ \cM$, if and only if $\cM = \alpha \, \id_\cV$ holds for some $\alpha \in \K$.
\end{lemma}
\begin{proof}
	This is a standard result of linear algebra. 
	We provide a proof for convenience of the reader. 
	Clearly, $\alpha \, \id_\cV$ commutes with all linear mappings~$\cM$ so we only need to prove the converse.

	Suppose that $\cM$ commutes with all isometries~$\cQ \in \cL(\cV,\cV)$. 
	For any given $w \neq 0$, we can decompose $\cM w$ as
	\begin{equation*}
		\cM w 
		= 
		\alpha \, w + z
		\quad 
		\text{where }
		z \perp w
		.
	\end{equation*}
	We choose an isometry $\cQ$ such that $\cQ \, w = w$ and $\cQ z = -z$ holds, \ie, a reflection at the $(n-1)$-dimensional subspace $z^\perp$.
	We compute
	\begin{equation*}
		\alpha \, w - z
		=
		\cQ (\alpha \, w + z)
		=
		\cQ \cM w
		=
		\cM \cQ \, w
		=
		\alpha \, w + z
		,
	\end{equation*}
	which implies $z = 0$. 
	Hence, every $w \in \cV$ is an eigenvector of $\cM$, implying that there is only one eigenspace of $\cM$ and thus $\cM = \alpha \, \id_\cV$ holds.
\end{proof}

With \cref{theorem:covariance-as-surface-integral} in place, we can derive the result that the metric trace of a sesquilinear form is, up to the dimension, the average of its quadratic form over the unit sphere:
\begin{theorem}
	\label{theorem:metric-trace-as-average-response-over-the-unit-sphere}
	Consider $\cV$ with an inner product $\inner{\cdot}{\cdot}_\cV$ and let $\sphere$ be the unit sphere \eqref{eq:inverse-covariance-unit-sphere-and-ball} with surface area~$\abs{\sphere}$.
	Then for any sesquilinear form~$a(\cdot,\cdot)$ on~$\cV$, we have
	\begin{equation}
		\label{eq:metric-trace-as-average-response-over-the-unit-sphere}
		\frac{1}{n} \trace_\cV a 
		= 
		\frac{1}{\abs{\sphere}} \int_{\sphere} a(v,v) \d v
		.
	\end{equation}
\end{theorem}
\begin{proof}
	We can interpret the expressions in \cref{cor:various-measures}  as expected values of $v \, v^\bidual$ for appropriate distributions, in our case for the uniform distribution on the unit sphere:
	\begin{equation*}
		\E(v \, v^\bidual)
		\coloneqq 
		\frac{1}{\abs{\sphere}}\int_{\sphere} v \, v^\bidual \d v 
		.
	\end{equation*}
	\Cref{theorem:covariance-as-surface-integral} now implies
	\begin{equation*}
		\cR_\cV^{-1}
		=
		n \, \E(vv^{**})
		.
	\end{equation*}
	Then by definition \eqref{eq:metric-trace} and \cref{theorem:metric-trace-of-sesquilinear-form-is-expected-value-of-quadratic-form} we obtain:
	\begin{equation*}
		\frac{1}{n} \trace_\cV a
		=
		\frac{1}{n} \conjugate{\trace \cR_\cV^{-1} \cA}
		=
		\E(a(v,v))
		=
		\frac{1}{\abs{\sphere}}\int_{\sphere} a(v,v) \d v
		.
	\end{equation*}
\end{proof}

As a special case of \cref{theorem:metric-trace-as-average-response-over-the-unit-sphere}, we obtain the following characterization of the classical trace of a square matrix~$A$:
\begin{corollary}
	In the case $\cV = \K^n$, endowed with the standard inner product, we obtain
	\begin{equation*}
		\frac{1}{n} \trace A
		=
		\frac{1}{\abs{S_{\K^n}}} \int_{S_{\K^n}} v^\transp A \, \conjugate{v} \d v
		.
	\end{equation*}
\end{corollary}

A second corollary of \cref{theorem:metric-trace-as-average-response-over-the-unit-sphere} arises from the fact that the Frobenius-type inner product and norm are defined in \eqref{eq:Frobenius-inner-product-for-linear-maps} using the metric trace. 

\begin{corollary}
	\label{corollary:Frobenius-inner-product-for-linear-maps-using-averages}
	Suppose that $\cS, \cT \in \cL(\cV,\cW)$ are linear maps between vector spaces $\cV$ and $\cW$ where $n = \dim \cV$.
	Both spaces are endowed with inner products $\inner{\cdot}{\cdot}_\cV$ and $\inner{\cdot}{\cdot}_\cW$, respectively.
	Then
	\begin{equation}
		\label{eq:Frobenius-inner-product-for-linear-maps-using-averages}
		\inner{\cS}{\cT}_{\cV \to \cW}
		=
		\frac{n}{\abs{\sphere}} \int_{\sphere} \inner{\cS v}{\cT v}_\cW \d v
	\end{equation}
	holds and in particular
	\begin{equation}
		\label{eq:Frobenius-norm-for-linear-maps-using-averages}
		\norm{\cT}_{\cV \to \cW}
		=
		\paren[auto](){\frac{n}{\abs{\sphere}} \int_{\sphere} \norm{\cT v}_\cW^2 \d v}^{1/2}
		.
	\end{equation}
\end{corollary}
As advertised, \cref{corollary:Frobenius-inner-product-for-linear-maps-using-averages} shows that the Frobenius norm has an interpretation similar to an operator norm.
Rather than the classical operator norm
\begin{equation*}
	\norm{\cT}_{\cL(\cV,\cW)}
	=
	\max_{0 \neq v \in \cV} \frac{\norm{\cT v}_\cW}{\norm{v}_\cV}
\end{equation*}
that measures the maximal $\cW$-norm response of~$\cT$ over the unit sphere in~$\cV$, the Frobenius-type norm \eqref{eq:Frobenius-norm-for-linear-maps-using-averages} takes into account the average norm response.

\begin{remark}
	As an alternative to \eqref{eq:metric-trace-as-average-response-over-the-unit-sphere}, we may evaluate $\trace_\cV a$ by integration over the unit ball~$\ball$ rather than the unit sphere~$\sphere$.
	In view of \cref{cor:various-measures}, we obtain
	\begin{equation*}
		\frac{1}{n} \trace_\cV a 
		=
		\frac{1}{\abs{\ball}} \int_{\ball} \frac{a(v,v)}{\inner{v}{v}_\cV} \d v
	\end{equation*}
	and, consequently,
	\begin{equation*}
		\norm{\cT}_{\cV \to \cW}
		=
		\paren[auto](){\frac{n}{\abs{\ball}} \int_{\ball} \frac{\norm{\cT v}_\cW^2}{\norm{v}_\cV^2} \d v}^{1/2}
		.
	\end{equation*}
	Similarly, we obtain from \cref{cor:various-measures} the following characterization, which is motivated by statistics:
	\begin{equation*}
		\frac1n \trace_\cV a 
		= 
		\frac{1}{(2\pi)^{n/2}} \int_\cV \frac{a(v,v)}{\norm{v}_\cV^2} \, \e^{- \frac{1}{2} \norm{v}_\cV^2} \d V.
	\end{equation*}
	or 
	\begin{equation*}
		\norm{\cT}_{\cV \to \cW}
		=
		\paren[auto](){\frac{n}{(2\pi)^{n/2}} \int_\cV \frac{\norm{\cT v}_\cW^2}{\norm{v}_\cV^2} \, \e^{- \frac{1}{2} \norm{v}_\cV^2} \d V}^{1/2}
		.
	\end{equation*}
\end{remark}

\section{Outlook on Preconditioned Gradient Methods for Neural Network Training}
\label{section:preconditioned-gradient-methods-for-neural-network-training}

Traditionally, the use of a non-standard inner product with respect to which the gradient of a function is evaluated is referred to as preconditioning.
In this section, we describe the use of Frobenius-type inner products in the minimization of an objective function depending on a matrix variable.
This situation is ubiquitous when training neural networks using gradient-based methods.
In this section, all vector spaces are over the field of \emph{real} numbers.

We keep the discussion generic and consider a differentiable, real-valued objective function $f \colon \Theta \to \R$ defined on some parameter space~$\Theta$ endowed with an inner product $\inner{\cdot}{\cdot}_\Theta$.
The direction of steepest descent is given by the unique solution of the strongly convex quadratic problem
\begin{equation}
	\label{eq:gradient-as-minimization-problem}
	\text{Minimize}
	\quad
	f(\theta)
	+
	f'(\theta) \, \delta \theta
	+
	\frac{1}{2} \inner{\delta \theta}{\delta \theta}_\Theta
	\quad
	\text{\wrt\ }
	\delta \theta \in \Theta
	,
\end{equation}
which amounts to
\begin{equation}
	\label{eq:steepest-descent-direction}
	\delta \theta
	=
	- \cR_\Theta^{-1} f'(\theta)
	\eqqcolon
	- \nabla_\Theta f(\theta)
	.
\end{equation}
In other words, the Riesz map~$\cR_\Theta$ associated with the inner product on~$\Theta$ acts as a preconditioner for a gradient method.

We can interpret the quadratic term $\inner{\delta \theta}{\delta \theta}_\Theta$ as a rough approximation of the remainder term of the first order model
\begin{equation*}
	f(\theta + \delta \theta)
	=
	f(\theta)
	+
	f'(\theta) \, \delta \theta
	+
	r_\theta(\delta \theta)
	.
\end{equation*}
If locally the remainder $r_\theta(\cdot) \approx \inner{\cdot}{\cdot}_\Theta$, we obtain a good fit between $f(\theta + \delta \theta)$ and its quadratic model, and thus the negative gradient \eqref{eq:steepest-descent-direction} provides a good step.
If, however, $r_\theta(\cdot)$ is very anisotropic with respect to $\norm{\cdot}_\Theta$, then the negative gradient $\delta \theta$ may point into a direction, where only very short steps achieve some decrease of~$f$. 
The main issue in choosing a preconditioner $\cR_\Theta$ is to find a choice that reflects $r_\theta$ reasonably well, while still allowing $\delta \theta$ to be evaluated inexpensively from \eqref{eq:steepest-descent-direction}.
For instance, if $f$~is convex, then $\cR_\Theta \coloneqq f''(x)$ is a choice that yields a good local but possibly expensive approximation of $r_\theta$. 

In neural networks for supervised learning tasks, the optimization variable $\theta = (\cS,b)$ usually has the structure of a linear mapping $\cS \in \cL(\cX,\cY)$ and a bias vector~$b \in \cY$, mapping input data $x \in \cX$ to labels $y = \cS x + b \in Y$. 
For further discussion, suppose that $(x,y) \in \cX \times \cY$ is a random variable due to a distribution $\d P(x,y)$, and independent samples $(x_i,y_i) \in \cX \times \cY$ for $i = 1, \dots, N$ from $(x,y)$ are available.

Gradient-based training in the simplest case of a single layer network can be described using the objective
\begin{equation*}
	f(\cS,b)
	\coloneqq
	\E_{x,y}[\ell(\cS x + b;y)]
	=
	\int_{\cX \times \cY} \ell(\cS x + b;y) \d P(x,y) 
	.
\end{equation*}
It is approximated by the empirical risk
\begin{equation*}
	f(\cS,b)
	\approx 
	\frac{1}{N} \sum_{i=1}^N \ell(\cS x_i + b;y_i)
	,
\end{equation*}
where $\ell(\cdot \, ;y) \colon \cY \to \R$ is a loss function.
Note that the second argument $y \in \cY$ after the semicolon is not an optimization variable.

As described in \eqref{eq:gradient-as-minimization-problem}, the gradient of~$f$ with respect to an inner product in the space of variables $\cL(\cX,\cY) \times \cY$ can be found from a quadratic optimization problem.
It is natural to consider only block-diagonal inner products, \ie,
\begin{equation*}
	\inner{\cS}{\cT}_{\cX \to \cY}
	+
	\inner{b}{c}_\cY
	,
\end{equation*}
which do not mix the linear (matrix) part and the bias vector offset.

Dropping the constant term in \eqref{eq:gradient-as-minimization-problem}, the quadratic problem yielding the gradient $\delta \theta = (\delta \cS, \delta b)$ is given by
\begin{equation*}
	\text{Minimize}
	\quad
	\E_{x,y}[\ell'(\cS x + b;y)(\delta \cS \, x,\delta b)]
	+
	\frac{1}{2} \inner{\delta \cS}{\delta \cS}_{\cX \to \cY}
	+
	\frac{1}{2} \inner{\delta b}{\delta b}_\cY
\end{equation*}
where minimization is \wrt $(\delta \cS,\delta b) \in \cL(\cX,\cY) \times \cY$.
By linearity of the derivative, and due to the assumed block-diagonal form of the inner product, this problem decouples into two separate subproblems for the gradient components $\delta \cS$ and $\delta b$:
\begin{subequations}
	\label{eq:gradient-of-loss-function}
	\begin{align}
		\text{Minimize}
		\quad
		&
		\E_{x,y}[\ell'(\cS x + b;y) \, \delta \cS \, x]
		+
		\frac{1}{2} \inner{\delta \cS}{\delta \cS}_{\cX \to \cY}
		\quad
		\text{\wrt\ }
		\delta \cS \in \cL(\cX,\cY)
		\label{eq:gradient-of-loss-function:1}
		\\
		\intertext{and}
		\text{Minimize}
		\quad
		&
		\E_{x,y}[\ell'(\cS x + b;y) \, \delta b]
		+
		\frac{1}{2} \inner{\delta b}{\delta b}_\cY
		\quad
		\text{\wrt\ }
		\delta b \in \cY
		.
		\label{eq:gradient-of-loss-function:2}
	\end{align}
\end{subequations}

Problem \eqref{eq:gradient-of-loss-function:2} is easy to solve and it yields
\begin{equation*}
	\delta b
	=
	- \cR_\cY^{-1} \, \E_{x,y}[\ell'(\cS x + b;y)]
	.
\end{equation*}
To cover \eqref{eq:gradient-of-loss-function:1}, we write 
\begin{equation*}
	\begin{aligned}
		\E_{x,y}[\ell'(\cS x + b;y) \, \delta \cS \, x]
		&
		=
		\trace \E_{x,y}[\ell'(\cS x + b;y) \, \delta \cS \, x]
		\\
		&
		=
		\E_{x,y}[\trace \ell'(\cS x + b;y) \, \delta \cS \, x]
		\\
		&
		=
		\E_{x,y}[\trace x \, \ell'(\cS x + b;y) \, \delta \cS]
		\\
		&
		=
		\trace \E_{x,y}[x \, \ell'(\cS x + b;y) \, \delta \cS]
		\\
		&
		\eqqcolon
		\trace (\cA \, \delta \cS)
		,
	\end{aligned}
\end{equation*}
where $\cA \coloneqq \E_{x,y}[x \, \ell'(\cS x + b;y)] \in \cL(\cY,\cX)$.
Plugging this into \eqref{eq:gradient-of-loss-function:1} and using a Frobenius-type inner product \eqref{eq:Frobenius-inner-product-for-linear-maps} allows to rewrite \eqref{eq:gradient-of-loss-function:1} as
\begin{equation*}
	\text{Minimize}
	\quad
	\trace (\cA \, \delta \cS)
	+
	\frac{1}{2} \cR_\cX^{-1} \delta \cS^* \cR_\cY \, \delta \cS
	\quad
	\text{\wrt\ }
	\delta \cS \in \cL(\cX,\cY)
	.
\end{equation*}
This problem has a unique solution $\delta \cS$ characterized by
\begin{equation}
	\label{eq:gradient-of-loss-function:problem}
	\trace (\cA \, \cH) 
	+ 
	\trace \cR_\cX^{-1} \delta \cS^* \cR_\cY \, \cH 
	=
	0
	\quad
	\text{for all } 
	\cH \in \cL(\cX,\cY)
	,
\end{equation}
or equivalently
\begin{equation}
	\label{eq:gradient-of-loss-function:optimality-condition:1}
	\cA
	+
	\cR_\cX^{-1} \delta \cS^* \cR_\cY 
	=
	0
	,
\end{equation}
which finally amounts to the negative gradient direction
\begin{equation}
	\label{eq:gradient-of-loss-function:optimality-condition:2}
	\delta \cS
	=
	-
	\cR_\cY^{-1} \cA^* \cR_\cX 
	.
\end{equation}
Depending on the dimensions of the spaces $\cX$ and $\cY$ and properties of $\cR_\cX$ and $\cR_\cY$ such as sparsity of other structure, the evaluation of the negative gradient by \eqref{eq:gradient-of-loss-function:optimality-condition:2} may be efficiently implementable.
Specifically, the linear system \eqref{eq:gradient-of-loss-function:optimality-condition:1} lends itself to an iterative solution, which will be described in more detail and in particular in the context of deep learning in a future publication.

As a special case we obtain the K-FAC (Kronecker-Factored Approximate Curvature) preconditioner \cite{MartensGrosse:2015:1}, which constructs $\cR_\cX$ and $\cR_\cY$.
The former is obtained by estimating the covariance of the data. 
Let us discuss reasonable generic choices for the norms in $\cX$ and $\cY$. 
For the space $\cX$ of input data, it seems natural to consider the inner product given by the inverse covariance of the data (also known as the Mahalanobis distance).
Given the input mean $\widehat{x} \coloneqq \E_x [x] = \int_\cX x \d P(x)$, which can be approximated by the empirical mean $\frac{1}{N} \sum_{i=1}^N x_i$, the data covariance is 
\begin{equation*}
	\cC
	\coloneqq
	\cov_x [x]
	=
	\cC \coloneqq \E_x [(x - \widehat{x})(x - \widehat{x})^\bidual] 
\end{equation*}
with empirical approximation $\frac{1}{N} \sum_{i=1}^N (x_i - \widehat{x})(x_i - \widehat{x})^\bidual$.
The Riesz map $\cR_\cX$ is then given by $\cR_\cX = \cC^{-1}$.
This means that perturbations in $\cX$ are measured relative to their covariance. 

Concerning the space~$\cY$, we may, for instance, use a second-order model
\begin{equation*}
	\cR_\cY 
	\coloneqq
	\E_{x,y} [\ell''(\cS x + b;y)]
	,
\end{equation*}
at least when $\ell$ is convex.
When $\ell$ is a score function, \ie, the logarithm of a parametrized distribution, this can be interpreted as a Fisher information matrix with respect to perturbations of the first argument. 
In that context there are some well known identities available:
\begin{equation*}
	\cR_\cY
	=
	\cov_{x,y} [\ell'(\cS x + b;y)]
	=
	\E_{x,y} [\ell'(\cS x + b;y) \, \ell'(\cS x + b;y)^\bidual]
	\in
	\cL(\cY,\cY^*)
	.
\end{equation*}
These are exactly the quantities, used by the K-FAC preconditioner, which is derived in \cite{MartensGrosse:2015:1} by simplifying the full Fisher information matrix.

\section{Conclusion}
\label{section:conclusion}

Our novel, general, and flexible definition of traces and Frobenius norms in the context of inner product spaces opens the door to new  research ideas.
For problems whose unknown is a linear mapping, our generalized norms can become a valuable tool for analysis, but also for the construction of new algorithms.
The interpretation of the Frobenius norm as an average action of a linear mapping indicates its usefulness in stochastic settings.
Thus, in particular in machine learning, new ideas may arise from this approach.
We plan to explore this avenue in future work.

% Insert the appendix.
\appendix

% Insert the bibliography.
\printbibliography

\end{document}